\documentclass[journal]{IEEEtran}

\usepackage{cite}
\usepackage[numbers,sort&compress]{natbib}

\ifCLASSINFOpdf
\else
\fi
\usepackage{graphicx} 
\usepackage{booktabs}
\usepackage{color}
\usepackage[justification=centering]{caption}
\usepackage{amsmath}
\usepackage{authblk}
\usepackage{blindtext}

\begin{document}

%
\title{Multi-Step Short-Term Wind Speed Prediction with Rank Pooling and Fast Fourier Transformation}

%
%
%

\author[1]{Hailong SHU}
\author[1]{Weiwei SONG} 
\author[1]{Zhen SONG}
\author[1]{Huichuang GUO} 
\author[1]{Chaoqun LI}  
\affil[1]{State Key Laboratory of NBC Protection for Civilian, Beijing 102205, China}

\affil[ ]{\textit {Corresponding Author: shl@pku.edu.cn}}
 

%



\maketitle

\begin{abstract}
Short-term wind speed prediction is essential for economical wind power utilization.
The real-world wind speed data is typically intermittent and fluctuating, presenting great challenges to existing shallow models.
In this paper, we present a novel deep hybrid model for multi-step wind speed prediction, namely LR-FFT-RP-MLP/LSTM (Linear Fast Fourier Transformation Rank Pooling Multiple-Layer Perception/Long Short-Term Memory).
Our hybrid model processes the local and global input features simultaneously.
We leverage Rank Pooling (RP) for the local feature extraction to capture the temporal structure while maintaining the temporal order.
Besides, to understand the wind periodic patterns, we exploit Fast Fourier Transformation (FFT) to extract global features and relevant frequency components in the wind speed data.
The resulting local and global features are respectively integrated with the original data and are fed into an MLP/LSTM layer for the initial wind speed predictions.
Finally, we leverage a linear regression layer to collaborate these initial predictions to produce the final wind speed prediction.
The proposed hybrid model is evaluated using real wind speed data collected from 2010 to 2020, demonstrating superior forecasting capabilities when compared to state-of-the-art single and hybrid models. Overall, this study presents a promising approach for improving the accuracy of wind speed forecasting.

\end{abstract}


%
\IEEEpeerreviewmaketitle

\section{Introduction}


\noindent With the exacerbation of global warming, the reduction of carbon emissions has become a shared objective for all nations. Utilizing renewable energy sources such as wind power, solar energy, and hydropower is one of the primary means of alleviating this issue \cite{jiang2018}. As a practical and forward-thinking form of green energy, wind power has made a significant contribution in recent years \cite{he2018}. For instance, the Global Wind Energy Council (GWEC) released its 2021 global wind report, which indicated a year-over-year (YoY) growth of over $53\%$ in the world wind power market in 2020 \cite{Lee2021}. Consequently, in order to effectively plan economically optimized load dispatch and efficiently utilize wind power, such as by properly adjusting wind power supplies, short-term wind speed prediction is critical. Furthermore, accurate predictions of short-term wind speeds can aid in advanced scheduling of cleaning, maintenance, and safety checks of wind turbines during periods of low wind conditions \cite{Shivam2020}. Due to the fact that wind speed can be affected by various factors and can therefore be quite unstable over time, accurately predicting wind speed remains an unresolved issue \cite{azimi2016}.


Over the past few decades, automatic wind speed prediction methods have been widely studied, which can be broadly categorized into two types: physical process-driven methods \cite{Zhang2019} and data-driven methods \cite{Yuan2017}. Physical process-driven methods rely on equations that describe the physical processes related to atmospheric changes over time \cite{kimura2002}, which simulate simplified atmospheric changes \cite{al2010}. In general, these equations can be derived through numerical weather prediction (NWP) \cite{Tian2020, James2018}, spatial correlation methods \cite{Ye2017}, computational fluid mechanics (CFD) methods \cite{Yang2018}, and the like. These methods usually generate high prediction accuracy and can be easily interpreted by humans. However, physical process-driven methods generally require collecting a large amount of meteorological data on the surface and upper air to establish complex mathematical physical models, making it time-consuming and impractical for short-term wind prediction \cite{Zhang2015}.



Alternatively, many recent studies have proposed data-driven methods that require fewer computational resources. Unlike physical process-driven methods, these methods learn the underlying hypothesis from historical wind speed data \cite{Qin2019} without requiring a large amount of meteorological data or prior physics knowledge for model construction. Data-driven methods can be further categorized into linear methods, nonlinear methods, hybrid methods \cite{shen2016forecasting}, and so on. Standard linear methods, including Moving Average (MA) \cite{Fernando2017}, Auto-regressive (AR) \cite{Liu2020}, Auto-regressive Moving Average (ARMA) \cite{Erdem2011}, and Auto-regressive Integrated Moving Average (ARIMA) models \cite{Cadenas2016}), explicitly explore the linear relationship in time series data. However, such methods may have poor generalization capability for data with nonlinear characteristics \cite{Zhang2016}. Therefore, many nonlinear models have also been employed to address this limitation, including Gaussian Process Regression (GPR) \cite{Cai2020}, Support Vector Regression (SVR) \cite{Chen2014}, Quantile Regression (QR) \cite{Nielsen2006}, Artificial Neural Networks (ANN) \cite{Li2010,mohandes1998neural}, and Long Short-Term Memory (LSTM) \cite{Gers2000,Liu2018}.


Moreover, to further improve prediction performance, some hybrid methods have also been proposed \cite{Kiplangat2016,chen2019two,wang2018multi}. Hybrid methods synthesize the advantages of single models and are typically composed of two or more methods, such as signal processing methods, statistical forecasting methods, and so on. One proposed hybrid methodology combines both linear and nonlinear modeling \cite{zhang2003time}. It is claimed that the hybrid model can improve forecasting accuracy achieved by either of the models if used separately \cite{hansen2003time}. This indicates that the hybrid approach has the potential to produce more accurate forecasts and exhibit better robustness.



In addition to the choice of modeling methods, another important aspect of wind speed prediction is the temporal modeling scheme, given that it is a time series prediction task. Brown \cite{brown1984}  proposed a temporal model for wind speed prediction, involving fitting autoregressive (AR) processes of various orders to hourly wind speed data and selecting the most suitable AR process using model selection criteria. The resulting distribution was approximately Gaussian and standardized to remove diurnal and seasonal nonstationarity by fitting a separate model. Such modeling was found to be useful in short-term operational decisions related to the integration of wind power into multiple-source energy systems. Milligan \cite{milligan2003statistical} applied ARIMA models to predict wind speed and power output up to one or six hours in advance and found that several alternative ARMA models performed well in forecasting over the testing time frame, but identifying the proper model was difficult in some cases. Torres et al. \cite{torres2005forecast} used the ARMA model to predict hourly averaged wind speed and found that the transformation and standardization of the time series were important for forming the appropriate model, with the ARMA model outperforming the persistence model. Costa et al. \cite{costa2008review} applied Kalman filters to predict wind speed and found that the persistence method performed better for hourly data, while the prediction model performed best for 5-minute time steps. However, due to the non-stationary, non-linear, and high-noise characteristics of historical wind speed data, it remains difficult to improve prediction accuracy using time series models  \cite{tao2018hierarchical}.



This paper aims to predict future wind speed by leveraging multi-scale temporal dependencies in meteorological data time series. To achieve this, we propose a spectral representation that decomposes the time series into multiple frequencies \cite{song2018human,song2020spectral}, each corresponding to a unique temporal scale. This spectral representation captures the global temporal information of the previous meteorological data, which is essential for accurate wind speed prediction. In addition, we extend the rank pooling algorithm \cite{Fernando2017} to specifically encode the temporal evolution of meteorological data, indicating the potential future trend of the data. To properly encode these temporal descriptors, we use the state-of-the-art MLP Mixture model \cite{tolstikhin2021mlp} as the regressor, which is known for its robustness and effectiveness, as well as its simplicity and lightweight design  \cite{benz2021adversarial}.
Our main contributions are summarized as follows:
\begin{itemize}
    
    \item  A deep learning time series prediction based on MLP/LSTM is introduced to explore and exploit the implicit information of wind speed time series for wind speed forecasting;
    
    \item  rank pooling \cite{fernando2015modeling} and fast Fourier transformation was extended to transform the wind speed series dataset for extracting high dimensional features ﬁrstly;
    
    \item Two diﬀerent models including rank pooling and fast Fourier transformation were applied to enhance the accuracy of the model by linear regression;
    
    \item The performance of the proposed EnsemMLP/ EnsemLSTM is successfully validated on 2 years data, to perform 1-hour ahead short-term wind speed forecasting. Statistical tests of experimental results have demonstrated the proposed EnsemMLP/ EnsemLSTM can achieve a satisfactory forecasting performance.
        
\end{itemize}



The remainder of this paper is arranged as follows: Section 2 introduces the methodology utilized in the proposed model, including MLP, LSTM, rank pooling, and fast Fourier transformation; Section 3 introduces the flowchart of the hybrid deep learning model for wind speed forecasting; Section 4 conducts the case study to prove the effectiveness of our model and compares the proposed model with some other involved models via performance evaluation indexes; Section 5 concludes the main discoveries of this study.

\section{Related Work}

\noindent In this section, we first systematically review previous wind speed prediction methods as well as their advantages and disadvantages in Sec.  \ref{subsec: wind-speed}. Since our task is a time series analysis, we also summarize current standard time series analysis methods in Sec. \ref{subsec: time series}.


\subsection{Wind speed prediction}
\label{subsec: wind-speed}

\noindent The prediction of wind speed can be divided into two main categories: physical-process-driven methods and data-driven methods. 

\textbf{The physical-process-driven methods} utilize mathematical descriptions of physical phenomena in the atmosphere and geographical conditions, such as primitive equations, momentum conservation, energy conservation, mass conservation, and water conservation equations, to predict wind speed \cite{zhang2019heuristic}. These methods have been widely explored in the past and can be seen as a problem of initial value \cite{jung2014current}. Assuming the initial atmospheric conditions are given, the equations for each atmospheric variable can be solved by applying physical forces over a period of time, resulting in the prediction of these variables within a given timeframe \cite{Wang2021review}. Numerical weather prediction (NWP) models have been widely used for large-scale and long-term weather predictions and have relatively stable accuracy. However, due to the high computational complexity of NWP models, they are not well-suited for short-term wind speed predictions. Data-driven methods are another approach to wind speed prediction that is based on statistical and machine learning models, which have become increasingly popular in recent years \cite{Tian2020}.


\textbf{Data-driven methods:} The prediction of wind speed can be achieved through data-driven methods that rely on statistical inference of data, requiring less prior knowledge of physical processes, and resulting in much lower computational costs \cite{fu2020composite}. These methods can be classified into three types: linear, nonlinear, and hybrid  \cite{shen2016forecasting}. 
Erdem et al. \cite{erdem2011arma} used autoregressive moving average (ARMA) and vector autoregression (VAR) methods to predict hourly mean wind attributes one hour ahead at two wind observation sites in North Dakota. Their study found that the VAR models offer a higher forecasting accuracy in wind direction and a closer performance in wind speed. 
Kavasser et al. \cite{kavasseri2009day} employed fractional-ARIMA models to forecast wind speed and power production for one and two days on the horizon, using their ability to incorporate long-range correlations in wind speed records. 
Liu et al. \cite{liu2011comprehensive}  utilized an autoregressive moving average-generalized autoregressive conditional heteroscedasticity (ARMA-GARCH) framework to model the mean and volatility of wind speed based on historical information, finding that the volatility of wind speed had a nonlinear and asymmetric time-varying property. 
Lastly, Zuluaga et al. \cite{zuluaga2015short} compared three different methods to make a Kalman filter robust to outliers in the context of one-step-ahead wind speed prediction and found that Robust Kalman filtering can serve as an accurate tool for electricity providers to predict wind speed.



Compared to linear methods, nonlinear methods can capture complex variations in wind speed using meteorological data  \cite{wu2015study}. For instance, Zhou et al. \cite{zhou2011fine} demonstrated that the support vector machine (SVM) outperformed the persistence model in one-step-ahead wind speed forecasting after fine-tuning its parameters. In a study in Zaragoza \cite{lahoz2006mlp}, a multi-layer perceptron (MLP) neural network was used to predict wind direction and speed. 
Iqdour et al. \cite{iqdour2006} also used MLP neural networks to predict actual wind speed and showed that the identified model could be successfully used for wind speed prediction. 
Mohandes et al. \cite{ mohandes2004support} applied SVM to wind speed prediction and compared its performance with that of multilayer perceptron neural networks. The prediction process using these methods was comparable to that of classical methods. 
Additionally, Vinothkumar et al. \cite{Vinothkumar2020} employed two machine learning models to forecast wind speed and found that the recurrent long short-term memory (LSTM) neural network model was more effective at predicting wind speed, while the wavelet SVM neural model was better at structural minimization principle and kernel function modeling, making it a better predictor.



Nonlinear methods have shown great potential in predicting wind speed variability based on meteorological data, especially in capturing complex and nonlinear relationships. However, these methods have some limitations, such as overfitting and local optima. 
To overcome these limitations, hybrid methods have been proposed that integrate various signal decomposition techniques, machine learning algorithms, and clustering approaches. These hybrid methods have shown improved prediction accuracy compared to single linear or nonlinear methods \cite{Sun2017, Qin2019,salcedo2009}.
The hybrid methods typically consist of two stages: signal decomposition and wind speed prediction \cite{zhao2019novel}. In the signal decomposition stage, the original wind speed time series is decomposed into several sub-sequences using various techniques. High-frequency subseries are usually ignored, while the remaining subseries are reconstructed to obtain stationary time series data. In the wind speed prediction stage, the new series data are used to predict wind speed using machine learning models or other prediction models.
In some studies, the decomposed subseries are used as input to a single prediction model, while in others, separate prediction models are assigned to each subseries. Finally, the subseries are composed to obtain the wind speed prediction results. This process can lead to better performance compared to single models, as it allows for taking advantage of individual techniques and can capture the underlying nonlinear relationships in the wind speed time series. 
Overall, hybrid methods have great potential in wind speed prediction and can contribute to the development of efficient and reliable wind energy systems.
 
In order to extract low-dimensional wind representation from the original data and remove noise, signal decomposition is often employed. Several signal processing algorithms, including Wavelet Analysis \cite{Zhang2016} and Fast Fourier Transformation \cite{zouaidia4}, Empirical Model Decomposition (EMD) \cite{Khosravi2018}, as well as other signal processing algorithms  \cite{Zheng2017, Jiajun2020, Tascikaraoglu2014} have been widely utilized for this purpose. Through these stages, relevant information such as time-frequency and trends can be effectively extracted. However, despite the benefits, these methods also suffer from certain limitations. For instance, wavelet-based methods are highly dependent on the decomposition level and wavelet basis, while the methods utilizing mode decomposition lack rigorous theoretical foundations. Moreover, other processing techniques such as EWT, VMD, and PSR may struggle to distinguish periodic components from quasi-periodic components in wind speed time series \cite{mi2020}. These limitations highlight the need for further research and innovation in signal processing algorithms to address these challenges and improve the accuracy of wind speed prediction.

The accurate selection of prediction methods is a crucial aspect of wind speed forecasting. A diverse range of prediction models has been implemented to predict wind speed by incorporating hybrid methods. For instance, the Kalman filter \cite{Gryning2000} was scrutinized to identify the optimal configuration for wind speed and wind power forecasting. The autoregressive moving average (ARMA) \cite{Zhang2016} method was employed to forecast wind speed and direction tuples. Additionally, a novel forecasting architecture based on AdaBoost neural networks in combination with wavelet decomposition was proposed to predict the wind speed in the short-term \cite{Kiplangat2016}. Furthermore, individual predictors such as long short-term memory (LSTM), extreme learning machine (ELM), and Elman neural network were selected to predict wind speed, and ELM-based combination mechanisms were utilized to obtain the final forecasts \cite{chen2019two}. 
Hu et al. \cite{hu2018nonlinear} utilized a differential evolutionary algorithm-optimized LSTM to accomplish nonlinear combinations and exhibit improved wind speed prediction results. 
In another study, Wang et al. \cite{wang2018multi} used the backpropagation (BP) neural network, Elman neural network, support vector machine (SVM), and autoregressive integrated moving average (ARIMA) to construct a hybrid model, where the BP, Elman, and ARIMA models were considered as the prediction part, and the SVM was treated as the ensemble part. The results indicate that the proposed prediction method possesses high accuracy and can effectively reflect the short-term wind speed patterns.


\begin{table*}[!ht]
\centering
\caption{The statistics of the collected dataset}
\begin{tabular}{ccccccccc}
\toprule  
&PRS& TEM& RHU& PRE1h& WD2mi& WS2mi& WD10mi& WS10mi\\
& (hPa) & (℃) & ($\%$) & (mm) & (°) & (m/s) & (°) & (m/s)\\
\midrule  
COUNT& 87558& 87558& 87558& 87558& 87558& 87558& 87558& 87558\\
MEAN& 995.78 & 6.21 & 55.25 & 0.05 & 200.40 & 2.73 & 201.24 & 2.73 \\
STD& 9.77 & 15.20 & 22.95 & 0.60 & 107.45 & 1.66 & 107.13 & 1.61 \\
MIN& 964.9& -34& 2& 0& 0& 0& 0& 0\\
$25\%$& 988.1& -7.2& 37& 0& 136& 1.5& 139& 1.6\\
$50\%$& 995.5& 8.2& 55& 0& 216& 2.4& 217& 2.4\\
$75\%$& 1003.4& 19.4& 73& 0& 292& 3.5& 292& .5\\
MAX& 1024.4& 42.1& 100& 40.6& 360& 14.9& 360& 13.8\\
\bottomrule 
\end{tabular}
\label{tab:Magin_settings}
\end{table*}

\subsection{time series temporal modelling}
\label{subsec: time series}

\noindent Time series models are commonly used in wind speed prediction; however, they rely on the assumption of data stationarity \cite{Mauricio1995}, which limits their prediction accuracy due to the strong nonlinearity of wind speed. These models include Moving Average (MA), Auto-regressive (AR) \cite{Liu2020}, Auto-regressive Moving Average (ARMA) \cite{Erdem2011} and their variants \cite{Cadenas2016}. To address the challenges of wind speed modeling, researchers have proposed methods that consider autocorrelation, non-Gaussian distribution and diurnal nonstationarity \cite{brown1984}. For instance, one study applied the Fourier transform to extract data information through preprocessing \cite{zouaidia4}. Another study developed a wind speed forecasting model using Fast Fourier Transform Filter with the Encoder-Decoder-LSTM model for 1- and 3-hour-ahead predictions \cite{zouaidia4}. In another study, Fourier analysis was combined with a nonlinear autoregressive network for wind speed prediction, using an open loop mode with target data generated by a Fourier model for multistep-ahead prediction \cite{rueda2021}. Recently, the rank pooling strategy has shown great success in summarizing short-term temporal evolution for various time series data analyses, such as action recognition \cite{Bilen2018}, facial emotion recognition \cite{song2019dynamic,song2021self}, activity recognition \cite{cherian2017generalized}, and personality recognition \cite{song2021person}.


\section{Data Collection}

\subsection{Collection Protocol}
\begin{figure}
\flushleft
\includegraphics[width=8cm]{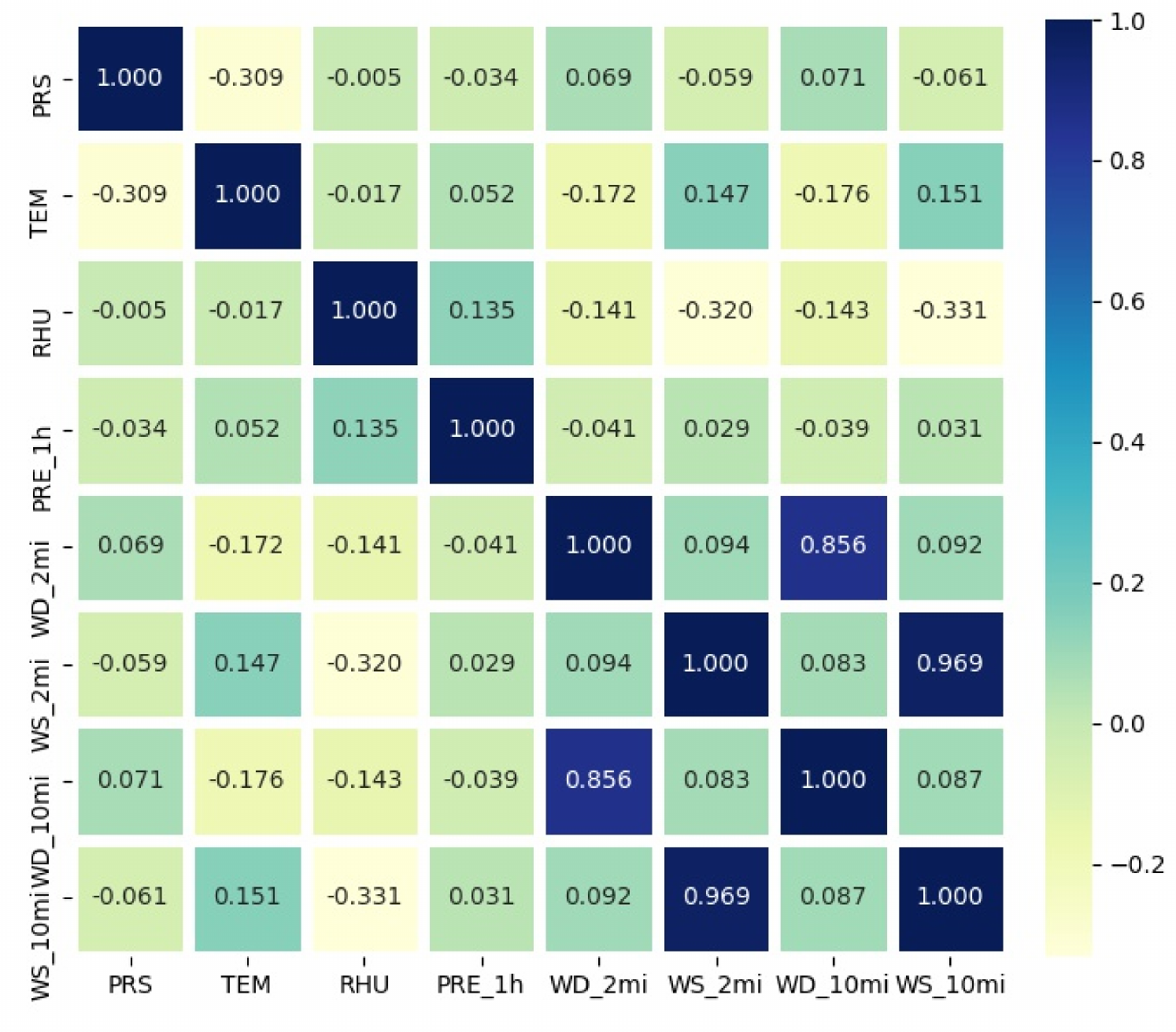} 
\caption{The correlation between each element} 
\end{figure}

\noindent In wind speed prediction studies, the temporal resolution of the wind speed data is an important factor to consider. Typically, wind speed data is collected at intervals of 10 minutes or 1-hour \cite{iqdour2006}. The 10-minute interval data is commonly utilized to forecast ultra-short-term wind speed, while the 1-hour interval data is often used to predict short-term wind speed \cite{Samadianfard2020}. Therefore, in order to investigate the performance of prediction models in the context of short-term wind speed forecasting, wind speed datasets with 1-hour intervals are frequently employed. This approach ensures that the selected models can accurately predict wind speed in a short-term horizon, providing valuable insights into the effectiveness of various prediction techniques.

\begin{figure*}[!ht]
\centering
\includegraphics[width=16cm]{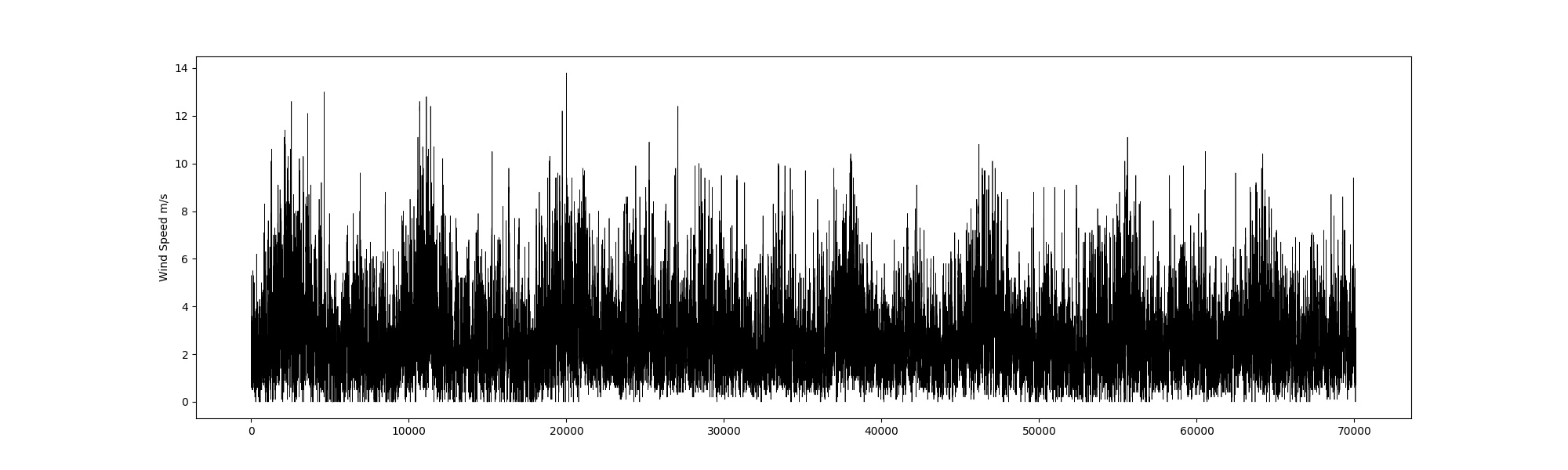} 
\centering
\caption{The original wind speed dataset} 
\end{figure*}
\subsection{Database Details}
\noindent This section presents a statistical exploration of the datasets used in this study. The surface meteorological data collected from a wind farm located in Inner Mongolia, China from 2010 to 2020 is analyzed, as depicted in Fig. 1. The dataset consists of 87558 samples with eight features, including pressure, temperature, humidity, precipitation, 2-minute average wind direction, 2-minute average wind speed, 10-minute average wind direction, and 10-minute average wind speed. The proposed model is evaluated by performing one-hour-ahead 10-minute average wind speed forecasting. The statistical characteristics of the dataset are summarized in Table \ref{tab:Magin_settings}. The range of wind speed values spans from 0 m/s to 13.8 m/s, with an average wind speed of 2.73 m/s and a median value of 2.4 m/s. The small difference between the mean and median values suggests a fairly balanced distribution of the data [50]. The hot map shown in Fig. 2 reveals that WS10MI exhibits the strongest correlation with WS2MI and a weak correlation with other variables.




\section{The Proposed Approach}
\noindent The dynamic nature of the atmosphere is characterized by a variety of parameters, including temperature, humidity, air pressure, and wind. In order to increase the accuracy of wind speed prediction, historical meteorological data collected by wind farms can be leveraged to create deep learning models. In this study, we propose a novel data-driven method that considers multi-scale, short-term temporal information of the atmosphere (as depicted in Fig. 3). Our approach begins by discussing temporal modeling and subsequently explores several algorithms that are integrated into our hybrid model, including rank pooling, MLP, LSTM, linear regression, and MLP-mixer.
\begin{figure}
\flushleft
\includegraphics[width=8cm]{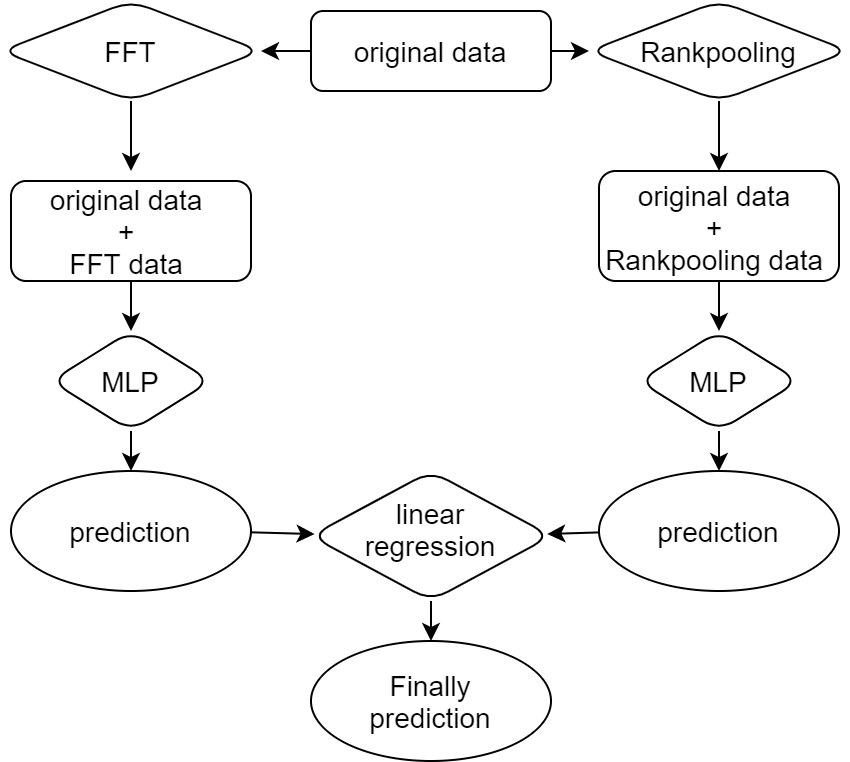} 
\caption{An artificial neural network arrangement in this study} 
\end{figure}

Notably, the rank pooling and mixer algorithms have been extensively utilized in the field of action recognition, and our proposed method integrates these contemporary algorithms with established methods to deliver more accurate wind speed forecasts. Our approach leverages the rank pooling and FFT methods to extract both recent and long-term prediction information from the data, and linear regression is then employed to combine these results in order to achieve the highest level of accuracy.


\subsection{time series Temporal Modelling}
Temporal modeling is a fundamental concept in signal processing that has been applied in diverse fields of scientific study. This technique facilitates the extraction of temporal information, investigation of dynamic properties, analysis of the correlation between past observations, and prediction of future events \cite{tao2018hierarchical}. For a long time, the Fast Fourier Transform (FFT) has been extensively utilized as a classical temporal model. However, with the advent of deep learning techniques, rank pooling has been gaining prominence in image and behavior recognition. In this study, we adopt both the FFT and rank pooling techniques to extract both long-term and local information from the wind speed data.

\subsubsection{Rank Pooling (wind speed trend modeling)}
Rank pooling is a learning-to-rank setup that computes a line in input space and projects input data onto this line, enabling pooling based on the temporal structure while preserving the temporal order \cite{ fernando2015modeling,fernando2016learning}. This technique has been extensively used in action recognition in video data by summarizing the video sequence based on the characteristics of the line. Rank pooling is an unsupervised learning-based temporal pooling technique that constructs a learning-to-rank model to gather relevant data from a composite activity's execution. New features of the composite activity are then created using the learned model's parameters. Rank pooling preserves the temporal ordering of the underlying activities and effectively functions even when there is no apparent temporal relationship between these sequences \cite{cherian2017generalized}. Recent studies have shown that rank pooling is a successful approach in identifying actions in video data \cite{Bilen2016}. In this study, we expand on this method to predict short-term wind speed by applying it to wind speed data and extracting relevant temporal features for accurate forecasting.

\subsubsection{Spectral Representation for Representing Multi-scale Short-term Temporal Information}
Temporal information is a critical factor in numerous scientific domains, and its appropriate treatment can yield valuable insights into the behavior of dynamic systems. The discrete Fourier transform (DFT) is a powerful mathematical tool that can convert specific sequences of functions into other types of representations, which can aid in the understanding of the underlying temporal structure of the data. The Fast Fourier Transformation (FFT) algorithm is commonly used to compute the DFT due to its efficiency and accuracy. Applications of FFT span across various fields, including engineering, music, science, and mathematics, and have demonstrated their effectiveness in data smoothing. In the present study, we employ the FFT to transform the time series data, as done in previous works. This transformation facilitates the calculation of the n-dimensional, n-point discrete Fourier transform via an efficient algorithm, which transforms the dataset from a time series domain to a smoother frequency domain. This transformation enables deep learning algorithms to process the data more efficiently and effectively.
\subsection{Prediction Model}
The accuracy and effectiveness of wind speed prediction greatly depend on the choice of the prediction model. In this section, we present a comprehensive discussion on the benchmarking models for wind speed prediction, as well as the proposed model. Firstly, we provide a concise introduction to Multilayer Perceptron (MLP), Long Short-Term Memory (LSTM), and MLP-mixer, which are widely used deep learning algorithms. Subsequently, we delve into the details of the proposed model, which combines the power of rank pooling and FFT methods to extract both short-term and long-term temporal information from historical meteorological data. The hybrid model also utilizes linear regression for integrating the two outcomes, leading to enhanced accuracy and precision in wind speed prediction.
\begin{figure}
\centering
\includegraphics[width=8cm]{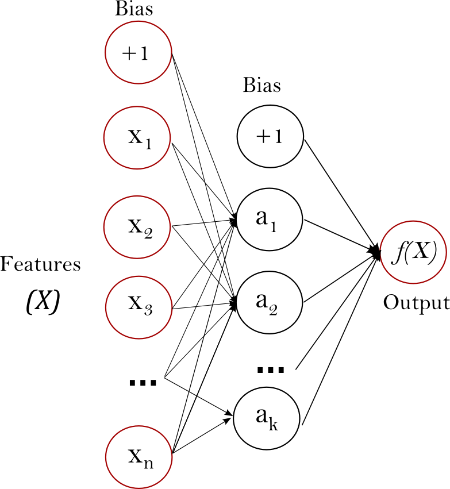} 
\caption{One hidden layer MLP} 
\end{figure}

\subsubsection{Long Short-term Memory (LSTM)}
Recurrent neural networks (RNNs) have been enhanced by the long short-term memory (LSTM) neural network, which enables feature learning beyond the drawback of gradient disappearance and has become a popular approach for wind speed forecasting. In this study, one of the prediction models employed is the LSTM. The LSTM network includes three gates: the forget gate, input gate, and output gate, which govern the memory of historical information in the network. Given a predicted wind speed time series $ y=(y1, y2, ...yt,...yT)$ and an input wind speed time series $x=( x1, x2, ...xt, ...xT) $ consisting of T time stamps, the LSTM calculation process is expressed as:
\begin{equation}
\begin{aligned}
ft &= sigmoid(Wf[ht—1, xt] + bf )\\
it &= sigmoid(Wi[ht—1, xt] + bi)	\\
C &= tanh(Wc[ht—1, xt] + bc)		\\
ot &= sigmoid(Wo[ht—1, xt] + bo)		\\
Ct &= f t*Ct—1 + it*C	\\
ht &= ot*softsign(Ct)		\\
yt &= sigmoid(Wy*ht + by)	\\  
\end{aligned}
\end{equation}
where $ft$, $it$ and $ot$ represent the forget gate, input gate and output gate \cite{Chen2021}, respectively. This calculation process is illustrated in Fig. 5.

In this work, a three-layered neural network structure comprising an input layer, a concealed layer, and an output layer is employed, as shown in Fig. 2. The input layer consists of eight independent parameters, including pressure, temperature, humidity, precipitation, 2-minute average wind direction, 2-minute average wind speed, 10-minute average wind direction, and 10-minute average wind speed, with wind speed serving as the dependent variable for the output. Each input layer has eight neurons, and each LSTM layer has four neurons and one dense layer. The activation function of the output layer is the sigmoid function, while the activation function of the LSTM layer is the hyperbolic tangent (tanh) function. The optimizer is Adam, the loss function is the mean squared error (MSE), and the EarlyStopping patience is set to 1500. The selection of these functions is determined by a process of trial and error to obtain precise wind speed estimates.

\subsubsection{Multi-layer Perceptron Method}
The multilayer perceptron (MLP) is a popular type of neural network model for predicting wind speed and is often selected as the approach of choice in artificial intelligence algorithms \cite{Zhang2016}. 
A function $ f(\cdot): R^m \rightarrow R^o$ is learned through training on a dataset where m is the number of input dimensions and o is the number of output dimensions. It is possible to learn a non-linear function approximator for either classification or regression given a set of features  $X = {x_1, x_2, ..., x_m}$ and a target y.
The MLP architecture consists of an input layer, one or more hidden layers, and an output layer. Each neuron in a given layer is fully connected to every neuron in the layer below it. The output of an MLP that is trained using backpropagation and an activation function in the output layer is a set of continuous values for prediction \cite{lahoz2006mlp}, with the loss function being the square error. Fig. 4 illustrates a single-hidden-layer MLP with scalar output.
In this study, the MLP was chosen as one of the prediction models for wind speed, and the ideal network design included an input layer with 8 neurons, a hidden layer with 3 levels (100, 200, 50 neurons), and an output layer with 8 neurons, the same as for the LSTM. The dependent variable and independent parameters are also identical to those of the LSTM. The activation function for the output layer was set to the sigmoid function, while the activation function for the hidden layers was set to the tanh function. The optimizer was set to Adam, the loss function was MSE, and the EarlyStopping patience was set to 1500. To ensure accurate wind speed estimates, the choice of these functions was based on a process of trial and error.


\subsubsection{MLP-mixer Method}
In recent years, the MLP-mixer has emerged as a highly efficient and succinct framework for information transportation across spatial characteristics. This model exploits the power of matrix translation and MLP to achieve global receptive fields \cite{yu2021rethinking}, thereby eliminating the need for the heavy attention module typically required by neural networks. MLP-mixer \cite{tolstikhin2021mlp} achieves this by projecting tokens and transposing matrices to capture long-range dependencies across image patches \cite{guo2021hire}. The channel-mixing and token-mixing MLPs are utilized to represent the relationship between tokens and channels. Notably, new architectures are continually being developed to further enhance the performance of MLP-based models \cite{guo2021hire}. 
In this study, the MLP-mixer is used, which is found to be effective in generating good experimental results using MLPs, skipping connection between layers, and normalizing layer \cite{lian2021mlp}. The three-layered structure utilized comprises an input layer, a hidden layer, and an output layer, with the dependent variable and independent parameters being identical to those used in the LSTM model. The ideal network design features 3 channels, 224 * 2 inputs, 1000 classes, 8 blocks, 16 patch size, 512 hidden units, 256 tokens MLP, and 2048 channels MLP.

\subsection{Proposed Hybrid Model}
In this study, we present a novel hybrid model for multi-step forecasting of wind speed that takes into account both the short- and long-term characteristics of wind speed series. The proposed model, named Local and Global Dynamic Representation Generation, combines the strengths of two powerful machine learning models, namely, multi-layer perceptron (MLP) and long short-term memory (LSTM) neural networks, in order to achieve high forecasting accuracy. The overall architecture of the proposed hybrid model is illustrated in Fig. 2, which comprises several key steps. 
Firstly, the rank pooling and fast Fourier transform (FFT) methods are employed to extract the distinctive features of the original wind speed data, resulting in two input datasets. 
These datasets are then integrated with the original data using two different procedures to create two input datasets. Next, the MLP and LSTM models are iteratively trained with the two input datasets until the loss value becomes stable, which is an indication that the model has converged. The initial prediction results are obtained from the trained models, and the final target prediction results are obtained by integrating the preliminary prediction findings using the linear regression method. Our proposed approach provides a novel and effective solution for the multi-step forecasting of wind speed that accounts for both short- and long-term characteristics of wind speed series.
\begin{figure}
\centering
\includegraphics[width=8cm]{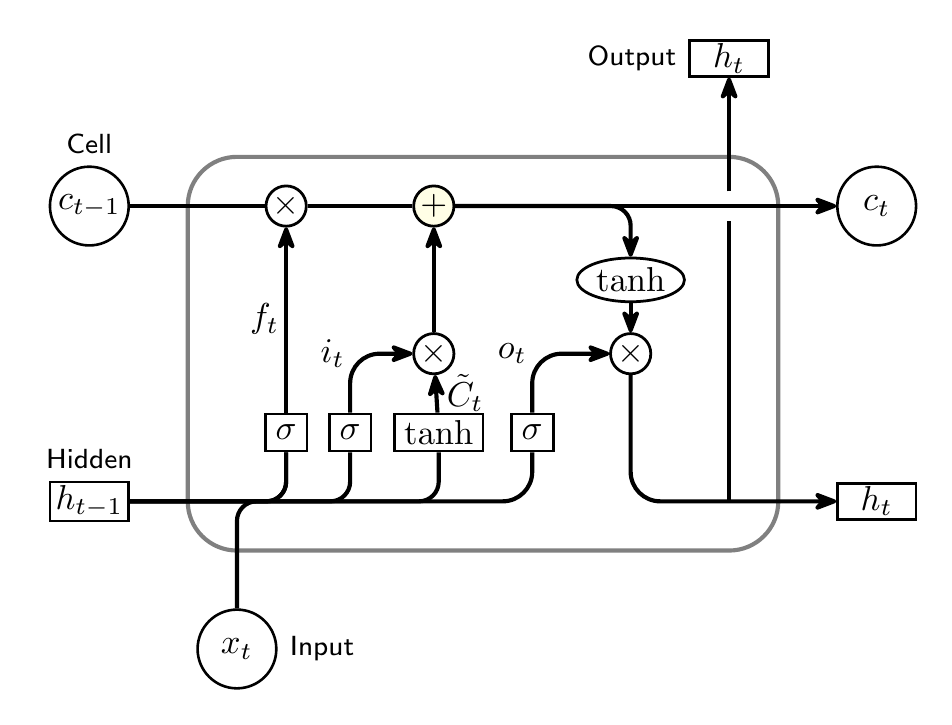} 
\caption{The architectural of LSTM network} 
\end{figure}



\subsubsection{Multi-scale Meteorological time series Local Dynamics Modelling}
This study proposes a novel approach for extracting local dynamics information from wind speed time series data. To this end, the rank pooling technique was used to process the original historical meteorological data in accordance with the z-score standard. $z = (x-\mu)/\sigma$,where $x$ is the raw score,$\mu$ is the population mean,and $\sigma$ is the population standard deviation. Specifically, the z-score was computed for matrix X using the mean and standard deviation for each column, with N, the length of the dimension that the z-score works along, being used to calculate the standard deviations. 

The resulting pre-processed dataset was then divided into groups based on the lookback time window, with retrospective 24-hour data from the predicted time being utilized to extract features using rank pooling for each sample taken 4 (8, 12, and 16) hours ago. This process allowed for the retention of more information in the previous meteorological data. The features of historical meteorological data were subsequently merged into 8*5 columns and coupled with the original meteorological data standardized between 0-1, forming the input dataset. 
Subsequently, the input dataset was used to train an MLP model, which underwent repeated training until the LOSS value was stable. Finally, the output of the test set was kept as the prediction results, and the model was evaluated using the test set data. The proposed approach can provide a reliable and efficient method for modeling local dynamics in wind speed time series data.

\subsubsection{Multi-scale Meteorological time series Global Dynamics Modelling}

In this study, the FFT method was utilized to extract the global characteristics of wind speed data, enabling time series Global Dynamics Modelling. The transformation was applied to extract the features of each dimension of the meteorological data, preserving the global information contained within the historical meteorological data. The modulus and complex angle of each meteorological data dimension were calculated and incorporated into 8*2 columns. Subsequently, the original meteorological data was combined with the features that the FFT extracted, producing an input dataset that was utilized to train the aforementioned MLP model iteratively. Similarly, the repetition was halted when 1500 consecutive LOSS values were larger than or equal to the preceding values. The output of the test set of the training set was stored as a prediction result, and the model was tested using the test set data. The proposed approach provides a novel and effective means to extract global characteristics from wind speed data and shows great potential for enhancing wind speed prediction accuracy.

\subsubsection{Local and Global Dynamic Representation Generation}

In this study, we propose a new hybrid model (Fig 3) for wind speed prediction that combines FFT and rank pooling to extract local and global information from the data, which is then integrated with the original data to produce the input dataset for the MLP model. The two MLP prediction results are combined using linear regression. 
To assess the performance of our proposed method, we compare it with seven other wind speed prediction techniques, including LSTM, MLP, rank pooling+LSTM, FFT+LSTM, rank pooling+MLP, MLP-mixer, etc. The wind speed dataset used is real wind speed data, and the performance of the prediction models is evaluated using MAE, RMSE, and R values, which are presented in Figures 6-8.
When using the LSTM approach as a prediction model, pre-processing of the historical weather data is required. This involves standardizing the historical data within a range of 0 to 1 before it can be input to the LSTM model. The rank pooling and FFT methods are employed to extract the features of the historical data, and the feature data and standardized historical data are then combined to form the input dataset. The output dataset generated by the two pre-processing techniques is used to iteratively train the LSTM model, with the repetition stopping when 1500 consecutive LOSS values are greater than or equal to the preceding values. The test dataset is used to evaluate the model, and the produced output results are saved as prediction results.
Finally, we also test a new prediction method called MLP-mixer for wind speed prediction. Our proposed hybrid model outperforms the other compared models in terms of prediction accuracy and robustness, highlighting the effectiveness of the proposed method for wind speed prediction.

\section{Experiments}




\noindent In this section, we evaluate the performance of the proposed hybrid model in predicting wind speed 6 hours ahead. We also compare the proposed model with other single or hybrid models to demonstrate its effectiveness and superiority. To this end, we constructed five different classification models for short-term wind speed prediction experiments. Specifically, we established two models for each of the five categories, for predicting wind speed 1-6 hours ahead. The aim of this analysis is to provide a comprehensive evaluation of the proposed hybrid model and its ability to outperform other state-of-the-art wind speed prediction models. 
The first classification serves as the control group and includes the MLP and LSTM models, with the design and parameter settings mentioned above. 
The second classification includes the application of fast Fourier transform (FFT-MLP, FFT-LSTM) to the MLP and LSTM models, where FFT is first used to extract the global feature information of the original meteorological data, and then deep learning approaches are employed to obtain the ultimate prediction results. 
The third classification includes the application of rank pooling (RP-MLP, RP-LSTM) to the MLP and LSTM models, where rank pooling is first used to extract the local feature information of the original meteorological data, and then deep learning approaches are used to obtain the ultimate prediction results. 
The fourth classification includes the application of fast Fourier transform through rank pooling (FFT-RP-MLP, FFT-RP-LSTM), where FFT and rank pooling are respectively used to extract the global and local feature information of the original meteorological data. The two kinds of feature data are then integrated as input data, and deep learning approaches are employed to obtain the ultimate prediction results. 
The fifth classification, the proposed approach, integrates FFT-MLP/ FFT-LSTM with RP-MLP/ RP-LSTM models using a linear regression (LR) model, named EnsemMLP/ EnsemLSTM. In this model, FFT and rank pooling are respectively used to extract the global and local feature information of the original meteorological data. The two kinds of feature data are then integrated as input data, and MLP/LSTM is used to obtain the preliminary prediction results. 
Finally, the preliminary results are input into the linear regression model to obtain the ultimate prediction result. Through these classifications, we are able to compare the performance of the proposed model against other single or hybrid models and demonstrate the effectiveness and superiority of our approach.

\begin{figure}[!ht]
\centering
\includegraphics[width=8cm]{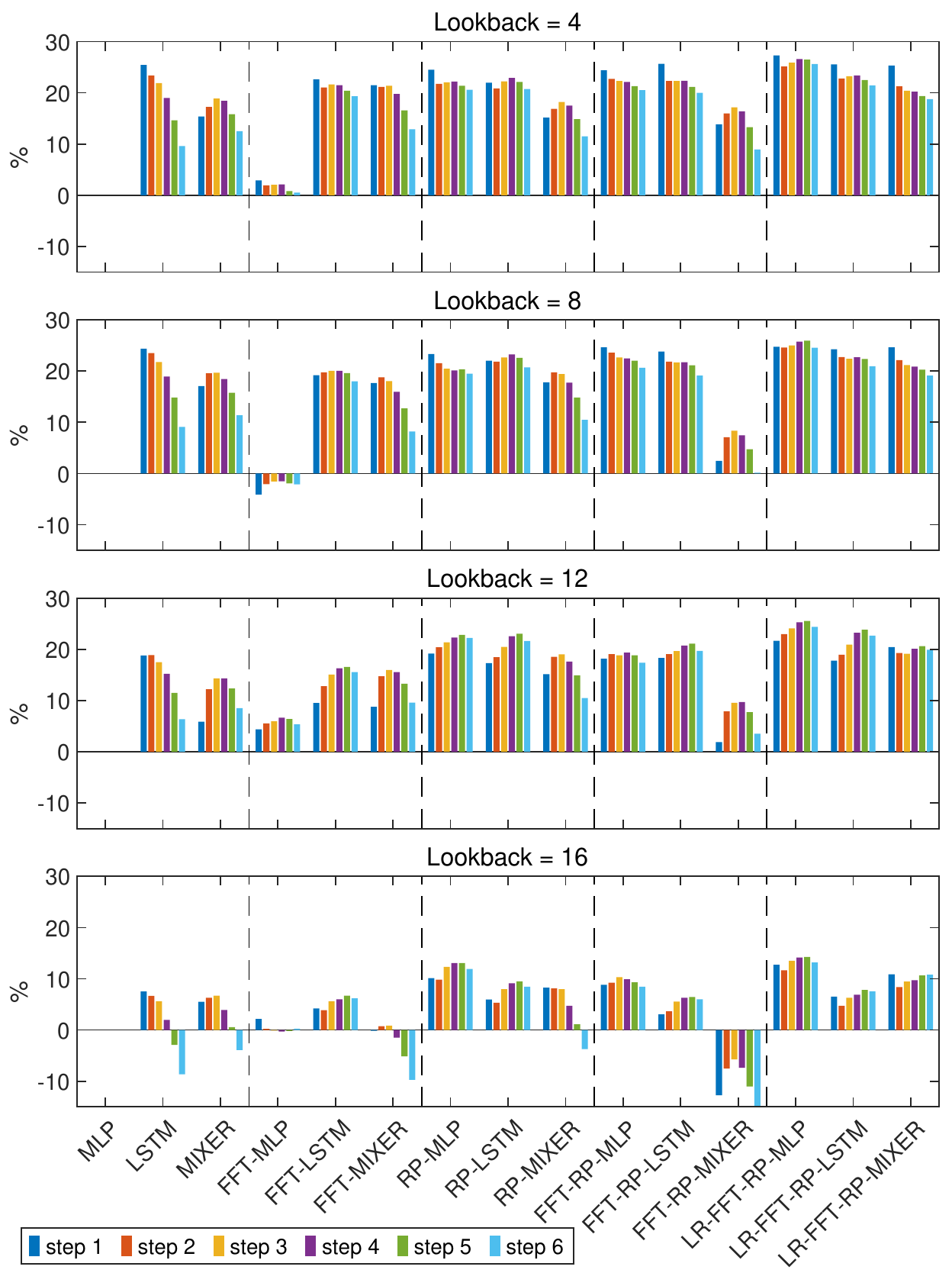}
\caption{
Percentage improvement of the mean absolute error (MAE) relative to MLP for each comparison model
} 
\end{figure}

\begin{figure}[!ht]
\centering
\includegraphics[width=8cm]{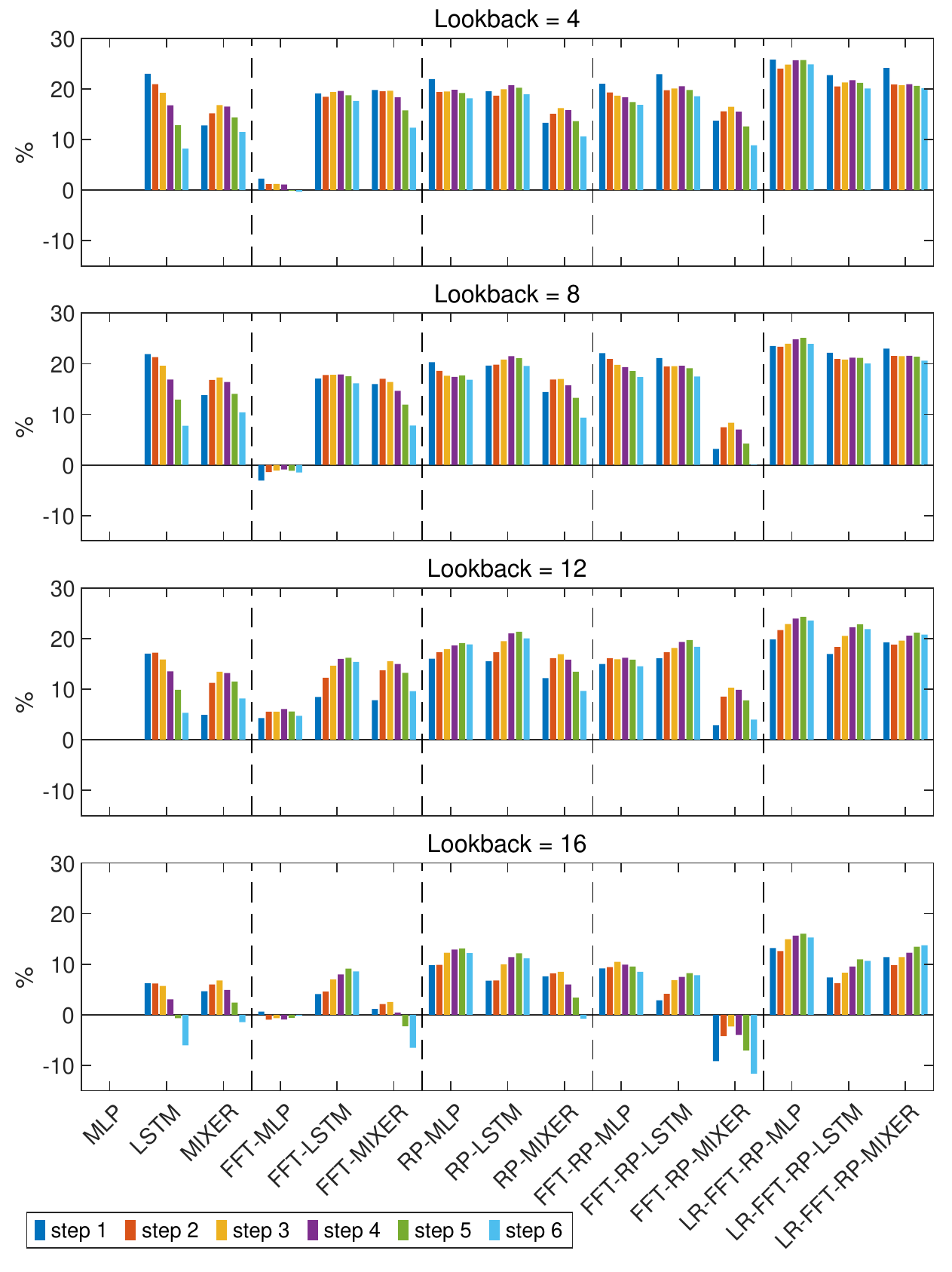}
\caption{
Percentage improvement of the root mean square error (RMSE) relative to MLP for each comparison model} 
\end{figure}

\begin{figure}[!ht]
\centering
\includegraphics[width=8cm]{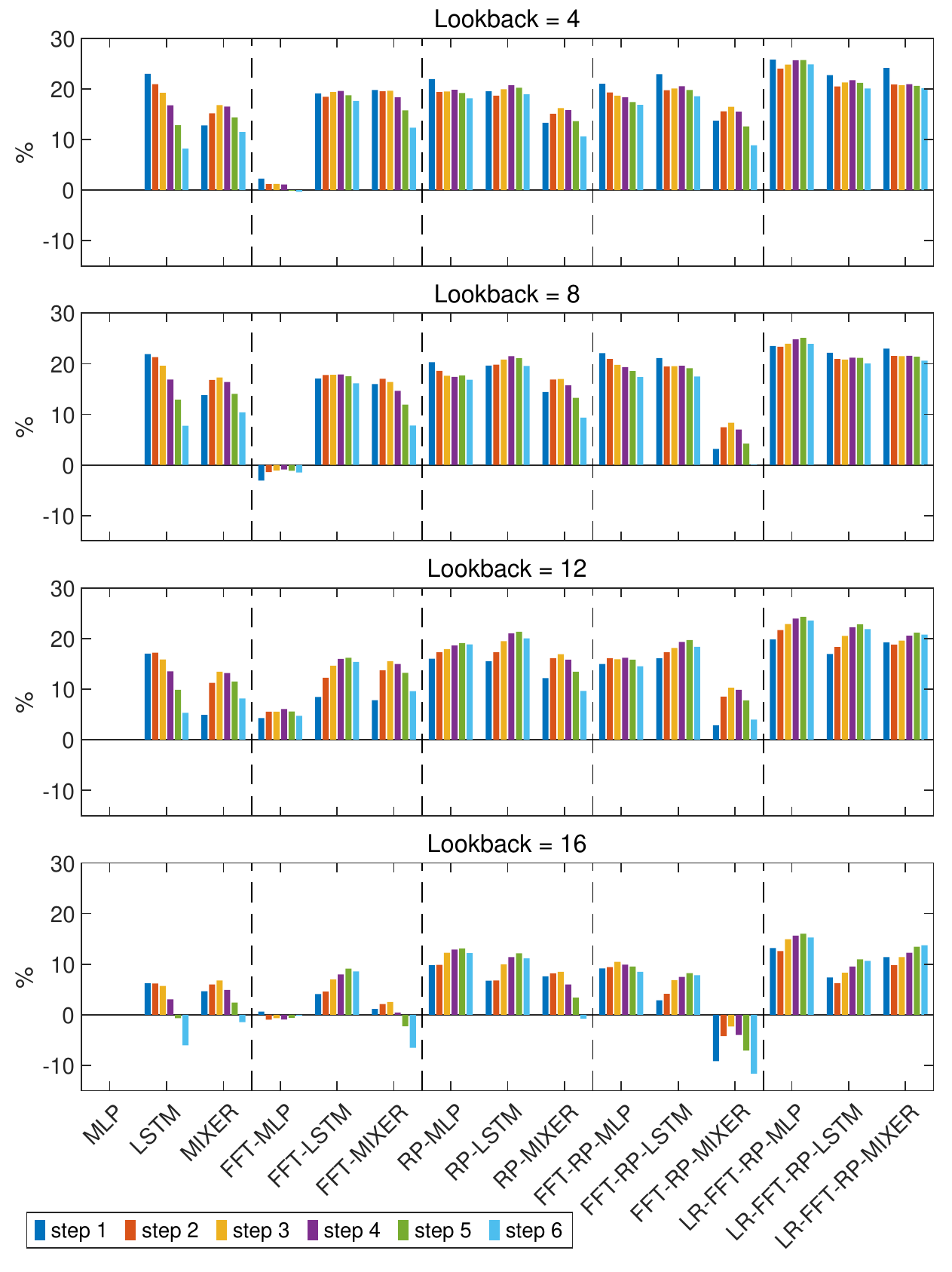} 
\caption{
Percentage improvement of the Pearson correlation coefficient (R) relative to MLP for each comparison model} 
\end{figure}

To assess the effectiveness and superiority of the proposed hybrid model in predicting wind speed, we compared its performance with those of several other single and hybrid models using mean average error (MAE), root mean square error (RMSE), and correlation coefficient (R) as evaluation metrics. The multistep MAE and RMSE results of different models are presented in Table II and Fig 6-8. Our findings show that the prediction error obtained by the proposed model is smaller than that obtained by other models. Specifically, the proposed model achieved the smallest prediction error and the best prediction performance among all models, as shown in Table II and Fig 6-7. These results demonstrate the efficacy and superiority of the proposed model in making effective wind speed predictions and obtaining a satisfactory data-fitting trend using real wind speed data.

\begin{table*}[htbp] 
\centering
\caption{Calculation results of error indexes of different models for multi-step I}
\begin{tabular}{cccccccccccccc}
\toprule  
&&\multicolumn{4}{c}{MAE}&
\multicolumn{4}{c}{RMSE}&
\multicolumn{4}{c}{R} \\
\midrule  
&Lookback&4&8&12&16&4&8&12&16&4&8&12&16\\
\midrule  
&Forecast &&&&&&&&&&&&\\
\midrule  
MLP
&1&0.865 &0.849 &0.798 &0.690 &1.153 &1.132 &1.074 &0.950 &0.763 &0.766 &0.786 &0.808  \\
&2&1.026 &1.017 &0.976 &0.824 &1.368 &1.357 &1.313 &1.136 &0.651 &0.652 &0.676 &0.719  \\
&3&1.170 &1.148 &1.109 &0.940 &1.556 &1.532 &1.494 &1.299 &0.537 &0.546 &0.570 &0.626  \\
&4&1.275 &1.247 &1.215 &1.018 &1.691 &1.660 &1.627 &1.406 &0.440 &0.456 &0.483 &0.557  \\
&5&1.333 &1.307 &1.280 &1.072 &1.767 &1.736 &1.709 &1.482 &0.378 &0.393 &0.417 &0.503  \\
&6&1.365 &1.328 &1.305 &1.100 &1.808 &1.768 &1.746 &1.520 &0.326 &0.343 &0.371 &0.468  \\
\midrule
LSTM
&1&0.645 &0.642 &0.648 &0.638 &0.888 &0.884 &0.891 &0.891 &0.845 &0.847 &0.844 &0.844  \\
&2&0.786 &0.778 &0.792 &0.769 &1.082 &1.068 &1.087 &1.066 &0.764 &0.771 &0.763 &0.771  \\
&3&0.914 &0.898 &0.915 &0.887 &1.257 &1.232 &1.257 &1.225 &0.678 &0.691 &0.680 &0.693  \\
&4&1.033 &1.011 &1.030 &0.998 &1.408 &1.380 &1.407 &1.363 &0.593 &0.609 &0.596 &0.616  \\
&5&1.138 &1.113 &1.133 &1.103 &1.541 &1.512 &1.540 &1.492 &0.510 &0.528 &0.514 &0.537  \\
&6&1.234 &1.207 &1.222 &1.195 &1.660 &1.631 &1.653 &1.612 &0.429 &0.449 &0.439 &0.457  \\
\midrule

FFT-MLP
&1&0.840 &0.884 &0.763 &0.675 &1.127 &1.167 &1.028 &0.944 &0.766 &0.767 &0.798 &0.805  \\
&2&1.006 &1.038 &0.922 &0.822 &1.352 &1.376 &1.240 &1.147 &0.645 &0.659 &0.702 &0.701  \\
&3&1.146 &1.166 &1.043 &0.941 &1.537 &1.549 &1.411 &1.308 &0.530 &0.553 &0.605 &0.607  \\
&4&1.248 &1.266 &1.134 &1.021 &1.672 &1.675 &1.528 &1.419 &0.434 &0.464 &0.529 &0.533  \\
&5&1.322 &1.332 &1.198 &1.074 &1.769 &1.756 &1.613 &1.491 &0.356 &0.400 &0.467 &0.478  \\
&6&1.358 &1.356 &1.235 &1.097 &1.814 &1.794 &1.663 &1.523 &0.302 &0.353 &0.418 &0.445  \\
\midrule

FFT-LSTM
&1&0.669 &0.686 &0.722 &0.661 &0.933 &0.939 &0.983 &0.911 &0.828 &0.826 &0.807 &0.837  \\
&2&0.810 &0.816 &0.851 &0.792 &1.116 &1.116 &1.152 &1.084 &0.741 &0.742 &0.722 &0.760  \\
&3&0.917 &0.918 &0.942 &0.887 &1.255 &1.259 &1.275 &1.208 &0.657 &0.657 &0.643 &0.690  \\
&4&1.001 &0.997 &1.017 &0.957 &1.360 &1.363 &1.367 &1.294 &0.579 &0.582 &0.573 &0.636  \\
&5&1.061 &1.051 &1.068 &1.000 &1.436 &1.432 &1.432 &1.347 &0.511 &0.524 &0.514 &0.597  \\
&6&1.101 &1.089 &1.102 &1.032 &1.490 &1.483 &1.477 &1.390 &0.454 &0.473 &0.467 &0.562  \\
\midrule

RP-MLP
&1&0.653 &0.651 &0.645 &0.620 &0.900 &0.902 &0.902 &0.857 &0.837 &0.830 &0.825 &0.845  \\
&2&0.803 &0.798 &0.777 &0.743 &1.103 &1.105 &1.086 &1.024 &0.748 &0.742 &0.740 &0.770  \\
&3&0.912 &0.913 &0.872 &0.824 &1.253 &1.262 &1.226 &1.140 &0.669 &0.663 &0.660 &0.705  \\
&4&0.992 &0.996 &0.944 &0.885 &1.356 &1.371 &1.324 &1.225 &0.606 &0.601 &0.592 &0.650  \\
&5&1.048 &1.041 &0.988 &0.932 &1.428 &1.429 &1.382 &1.288 &0.560 &0.558 &0.545 &0.604  \\
&6&1.084 &1.069 &1.015 &0.969 &1.480 &1.470 &1.417 &1.335 &0.515 &0.515 &0.509 &0.563  \\
\midrule

RP-LSTM
&1&0.675 &0.662 &0.660 &0.649 &0.928 &0.910 &0.907 &0.886 &0.838 &0.840 &0.840 &0.846  \\
&2&0.812 &0.795 &0.796 &0.780 &1.113 &1.088 &1.086 &1.059 &0.754 &0.762 &0.762 &0.771  \\
&3&0.910 &0.888 &0.882 &0.865 &1.246 &1.213 &1.203 &1.170 &0.679 &0.693 &0.699 &0.711  \\
&4&0.983 &0.957 &0.941 &0.925 &1.341 &1.303 &1.285 &1.246 &0.614 &0.635 &0.651 &0.664  \\
&5&1.038 &1.012 &0.985 &0.970 &1.410 &1.370 &1.344 &1.302 &0.559 &0.586 &0.612 &0.626  \\
&6&1.082 &1.053 &1.023 &1.007 &1.466 &1.422 &1.396 &1.351 &0.507 &0.544 &0.574 &0.589  \\
\midrule

FFT-RP-MLP
&1&0.654 &0.640 &0.653 &0.629 &0.911 &0.882 &0.913 &0.863 &0.820 &0.830 &0.823 &0.839  \\
&2&0.793 &0.777 &0.790 &0.748 &1.104 &1.073 &1.101 &1.029 &0.729 &0.743 &0.739 &0.768  \\
&3&0.909 &0.888 &0.900 &0.843 &1.266 &1.229 &1.256 &1.163 &0.641 &0.658 &0.660 &0.703  \\
&4&0.993 &0.967 &0.980 &0.917 &1.381 &1.339 &1.363 &1.267 &0.573 &0.593 &0.601 &0.650  \\
&5&1.049 &1.019 &1.039 &0.972 &1.460 &1.414 &1.438 &1.341 &0.519 &0.546 &0.554 &0.607  \\
&6&1.085 &1.054 &1.078 &1.007 &1.504 &1.461 &1.492 &1.391 &0.480 &0.511 &0.509 &0.573  \\
\midrule

FFT-RP-LSTM

&1&0.643 &0.647 &0.652 &0.669 &0.889 &0.893 &0.901 &0.923 &0.845 &0.843 &0.841 &0.832  \\
&2&0.797 &0.795 &0.790 &0.794 &1.098 &1.093 &1.086 &1.089 &0.751 &0.754 &0.759 &0.757  \\
&3&0.909 &0.899 &0.891 &0.888 &1.244 &1.233 &1.223 &1.210 &0.666 &0.674 &0.681 &0.689  \\
&4&0.990 &0.976 &0.963 &0.954 &1.344 &1.335 &1.312 &1.301 &0.592 &0.602 &0.620 &0.629  \\
&5&1.051 &1.031 &1.010 &1.003 &1.418 &1.404 &1.372 &1.360 &0.527 &0.545 &0.574 &0.586  \\
&6&1.092 &1.074 &1.048 &1.034 &1.473 &1.459 &1.425 &1.401 &0.471 &0.493 &0.527 &0.551  \\
\midrule

LR-FFT-RP-MLP

&1&0.629 &0.639 &0.625 &0.602 &0.856 &0.866 &0.861 &0.825 &0.837 &0.833 &0.835 &0.850  \\
&2&0.768 &0.767 &0.752 &0.728 &1.040 &1.040 &1.028 &0.993 &0.748 &0.747 &0.754 &0.773  \\
&3&0.867 &0.861 &0.842 &0.813 &1.170 &1.165 &1.152 &1.105 &0.668 &0.669 &0.678 &0.708  \\
&4&0.936 &0.926 &0.908 &0.874 &1.257 &1.248 &1.237 &1.186 &0.602 &0.606 &0.615 &0.653  \\
&5&0.980 &0.968 &0.953 &0.919 &1.313 &1.300 &1.294 &1.245 &0.553 &0.562 &0.567 &0.608  \\
&6&1.015 &1.002 &0.987 &0.955 &1.359 &1.345 &1.335 &1.288 &0.508 &0.518 &0.527 &0.570  \\
\midrule

LR-FFT-RP-LSTM

&1&0.644 &0.643 &0.656 &0.645 &0.891 &0.881 &0.892 &0.880 &0.844 &0.848 &0.844 &0.851  \\
&2&0.792 &0.786 &0.791 &0.785 &1.088 &1.073 &1.072 &1.065 &0.756 &0.764 &0.764 &0.770  \\
&3&0.898 &0.891 &0.877 &0.881 &1.225 &1.213 &1.187 &1.191 &0.677 &0.685 &0.701 &0.702  \\
&4&0.977 &0.964 &0.933 &0.948 &1.324 &1.308 &1.265 &1.272 &0.607 &0.621 &0.653 &0.652  \\
&5&1.033 &1.015 &0.975 &0.988 &1.393 &1.369 &1.319 &1.320 &0.550 &0.574 &0.616 &0.618  \\
&6&1.072 &1.050 &1.009 &1.017 &1.445 &1.414 &1.364 &1.358 &0.501 &0.534 &0.581 &0.587  \\

\bottomrule 
\end{tabular}
\end{table*}

\begin{table*}[htbp] 
\centering
\caption{Calculation results of error indexes of models based on MIXER for multi-step II}
\begin{tabular}{cccccccccccccc}
\toprule  
&&\multicolumn{4}{c}{MAE}&
\multicolumn{4}{c}{RMSE}&
\multicolumn{4}{c}{R} \\
\midrule  
&Lookback&4&8&12&16&4&8&12&16&4&8&12&16\\
\midrule  
&Forecast &&&&&&&&&&&&\\
\midrule  
MIXER
&1&0.732 &0.704 &0.751 &0.652 &1.006 &0.976 &1.021 &0.906 &0.817 &0.825 &0.823 &0.829  \\
&2&0.849 &0.818 &0.857 &0.772 &1.161 &1.129 &1.165 &1.068 &0.731 &0.739 &0.739 &0.746  \\
&3&0.949 &0.922 &0.950 &0.877 &1.295 &1.267 &1.293 &1.211 &0.645 &0.652 &0.655 &0.662  \\
&4&1.040 &1.017 &1.041 &0.978 &1.412 &1.388 &1.412 &1.337 &0.563 &0.567 &0.570 &0.578  \\
&5&1.122 &1.101 &1.122 &1.066 &1.513 &1.492 &1.512 &1.446 &0.487 &0.488 &0.493 &0.500  \\
&6&1.194 &1.177 &1.194 &1.143 &1.601 &1.584 &1.603 &1.542 &0.416 &0.413 &0.417 &0.426  \\
\midrule

FFT-MIXER
&1&0.679 &0.699 &0.728 &0.691 &0.925 &0.951 &0.990 &0.939 &0.823 &0.821 &0.785 &0.823  \\
&2&0.809 &0.826 &0.832 &0.818 &1.101 &1.126 &1.133 &1.112 &0.738 &0.738 &0.709 &0.743  \\
&3&0.920 &0.941 &0.932 &0.932 &1.251 &1.281 &1.262 &1.266 &0.653 &0.651 &0.633 &0.661  \\
&4&1.023 &1.048 &1.026 &1.033 &1.381 &1.417 &1.383 &1.400 &0.570 &0.567 &0.553 &0.580  \\
&5&1.112 &1.141 &1.110 &1.127 &1.489 &1.529 &1.483 &1.516 &0.496 &0.490 &0.482 &0.504  \\
&6&1.189 &1.219 &1.180 &1.207 &1.585 &1.630 &1.578 &1.619 &0.425 &0.417 &0.409 &0.432  \\
\midrule
RP-MIXER
&1&0.734 &0.698 &0.677 &0.633 &1.000 &0.969 &0.943 &0.878 &0.822 &0.822 &0.826 &0.837  \\
&2&0.853 &0.816 &0.795 &0.757 &1.162 &1.128 &1.101 &1.043 &0.734 &0.737 &0.744 &0.755  \\
&3&0.957 &0.925 &0.898 &0.865 &1.304 &1.272 &1.241 &1.189 &0.647 &0.650 &0.661 &0.671  \\
&4&1.052 &1.026 &1.001 &0.970 &1.424 &1.399 &1.369 &1.322 &0.565 &0.564 &0.576 &0.585  \\
&5&1.135 &1.113 &1.089 &1.060 &1.526 &1.506 &1.479 &1.432 &0.490 &0.486 &0.497 &0.506  \\
&6&1.208 &1.189 &1.168 &1.141 &1.616 &1.603 &1.577 &1.532 &0.419 &0.409 &0.421 &0.429  \\
\midrule

FFT-RP-MIXER

&1&0.745 &0.828 &0.783 &0.778 &0.995 &1.096 &1.043 &1.037 &0.791 &0.772 &0.777 &0.785  \\
&2&0.862 &0.945 &0.899 &0.886 &1.155 &1.256 &1.201 &1.184 &0.718 &0.699 &0.701 &0.719  \\
&3&0.969 &1.052 &1.003 &0.994 &1.300 &1.404 &1.340 &1.329 &0.642 &0.623 &0.625 &0.646  \\
&4&1.066 &1.154 &1.097 &1.093 &1.429 &1.544 &1.466 &1.462 &0.568 &0.543 &0.549 &0.571  \\
&5&1.156 &1.245 &1.181 &1.190 &1.545 &1.663 &1.576 &1.587 &0.494 &0.469 &0.477 &0.494  \\
&6&1.243 &1.326 &1.259 &1.276 &1.648 &1.766 &1.676 &1.697 &0.425 &0.401 &0.407 &0.421  \\
\midrule

LR-FFT-RP-MIXER

&1&0.646 &0.640 &0.635 &0.615 &0.875 &0.872 &0.867 &0.842 &0.829 &0.830 &0.832 &0.843  \\
&2&0.808 &0.792 &0.788 &0.755 &1.083 &1.065 &1.066 &1.025 &0.721 &0.733 &0.732 &0.755  \\
&3&0.931 &0.905 &0.897 &0.851 &1.234 &1.203 &1.201 &1.151 &0.614 &0.640 &0.642 &0.679  \\
&4&1.017 &0.987 &0.971 &0.919 &1.337 &1.302 &1.292 &1.234 &0.519 &0.555 &0.567 &0.619  \\
&5&1.075 &1.042 &1.016 &0.958 &1.403 &1.365 &1.347 &1.283 &0.442 &0.491 &0.513 &0.578  \\
&6&1.109 &1.074 &1.045 &0.981 &1.444 &1.404 &1.383 &1.311 &0.384 &0.444 &0.475 &0.551  \\

\bottomrule 
\end{tabular}
\end{table*}

To provide a more detailed analysis of the prediction results obtained through different methods, we refer to Table II and Fig 6-8. Based on these results, we can draw some important conclusions as follows:

(1) For the control model, which used original meteorological data to predict short-term wind speed, LSTM exhibited a superior predicting ability compared to MLP in terms of MAE, RMSE, and R for every prediction lookback window. When the original meteorological data was pre-processed through rank pooling, the prediction result of MLP was significantly improved, while that of LSTM showed limited improvement. This is due to the fact that MLP has a larger proportion of data information at the nearby time of prediction, while LSTM contains a long-term memory algorithm. Therefore, the different time scale features extracted through rank pooling can compensate for the deficiencies of MLP but do little to improve the prediction result of LSTM. When the original meteorological data were pre-processed by FFT, there was little improvement in the prediction result of MLP or LSTM. In the first classification, FFT had a small contribution to improving the prediction results for both prediction models, whether using FFT alone to process the data or in combination with the RP method.

(2) The LSTM model showed good prediction results for steps 1-2, with MAE (RMSE) below 0.8m/s (1.1m/s) for different lookbacks (Table II). However, except for the proposed model, the improvement of LSTM was not significant (Fig 6,7). As the number of prediction steps increased, the RMSE of various models exhibited an ascending behavior, while R showed a descending behavior. As the number of prediction lookback windows increased, the RMSE of each model did not show an obvious change trend, except for the proposed model. As can be observed from the RMSE in steps 3-6 of the proposed model (Fig.7), the RMSE exhibited a descending behavior with an increase in prediction lookback windows.

(3) Comparing the second classification (FFT-MLP, FFT-LSTM) with control models, we found that the second classification had some improvement for the control model, with the FFT-LSTM model exhibiting significant improvement for the prediction results of steps 4-6 of LSTM, but not for steps 1-3. FFT-MLP showed little improvement for the MLP model. When the lookback was less than 12, the RMSE of MLP and MLP-FFT was significantly larger than that of control models, while the RMSE of FFT-LSTM became larger than that of LSTM. For the MLP model, the RMSE decreased slightly when the lookback was 4 or 12, but became larger when the lookback was 8.

(4) A comparative analysis of the third classification model, which includes RP-MLP and RP-LSTM models, with the control models, clearly demonstrates the effectiveness of rank pooling in improving the precision of multistep wind speed prediction. The mean absolute error (MEA) and root mean square error (RMSE) of the third classification model were significantly lower than those of the control models. Furthermore, a comparison of the RP-LSTM model with two congeneric models, i.e., LSTM and FFT-LSTM, revealed that the third classification model outperformed the other two models. This demonstrates the advantages of LSTM in wind speed prediction. In addition, comparing the RP-LSTM model with LSTM, we found that the MEA of the latter was higher than that of the former, while the RMSE of each lookback for the LSTM model was slightly reduced. For the MLP model, the RMSE of each lookback was significantly reduced, and in fact, when the lookback was 12 or 16, the MLP model even outperformed the LSTM model in terms of predictive accuracy after RP preprocessing. These findings validate the efficacy of rank pooling in multistep wind speed prediction and highlight its potential for improving prediction accuracy.
(5) The fourth classification model, which includes FFT-RP-LSTM and FFT-RP-MLP models, showed no significant improvement in the prediction results of LSTM and MLP models after FFT and RP pre-processing. When the lookback was 4 or 8, the prediction results of the fourth classification model were similar to those of the third classification model. However, when the lookback was 12 or 16, the prediction error of the fourth classification model increased. This indicates that while FFT and RP pre-processing may improve prediction accuracy in some cases, it may not always lead to significant improvements, and its effectiveness may depend on the specific conditions of the wind speed prediction problem.
(6) The fifth classification model, i.e., the proposed model, which includes LR-FFT-RP-LSTM and LR-FFT-RP-MLP models, showed the greatest improvement in wind speed prediction compared to the previous classification models. The LR-FFT-RP-LSTM model had the greatest improvement when the lookback was 12, and the LR-FFT-RP-MLP model had the greatest improvement when the lookback was 16. Overall, the LR-FFT-RP-MLP model outperformed the LR-FFT-RP-LSTM model in terms of wind speed prediction accuracy. Specifically, for the results predicted in step 1, the MAE of the LR-FFT-RP-MLP model was reduced from 12.8$\%$ to 27.3$\%$ compared to MLP, and the RMSE was reduced from 13.2$\%$ to 25.8$\%$ compared to MLP. For the results predicted in step 6, the MAE of the LR-FFT-RP-MLP model was reduced from 13.2$\%$ to 25.6$\%$ compared to MLP, and the RMSE was reduced from 15.3$\%$ to 24.8$\%$ compared to MLP. These results suggest that the proposed LR-FFT-RP-MLP model is highly effective in improving wind speed prediction accuracy, especially for long-term predictions.

(7) Finally, to further evaluate the proposed method, the MLP-Mixer approach was used as a benchmark comparison. Similar preprocessing steps, including rank pooling and FFT, were applied to the MLP-Mixer method. The results, as shown in Figures 7-8, indicate that without pre-processing, the predicted improvement of each step is comparable to that of MLP. Unlike LSTM, the Mixer method exhibited the greatest improvement for steps 3-4, while the decrease in prediction accuracy for step 6 was not as significant as LSTM.
When the lookback was set to 16, the FFT-MIXER and FFT-RP-MIXER models exhibited worse prediction results than MLP. When the lookback was set to 4, 8, or 12, the RP-MIXER and FFT-RP-MIXER models had worse prediction improvements than RP-MLP, FFT-RP-MLP, RP-LSTM, and FFT-RP-LSTM. However, the FFT-MIXER and LR-FFT-RP-MIXER models with FFT preprocessing had comparable prediction results with FFT-MLP, LR-FFT-RP-MLP, FFT-LSTM, and LR-FFT-RP-LSTM.
In summary, the results demonstrate the effectiveness of the proposed LR-FFT-RP-MLP model for multistep wind speed prediction. The proposed method outperformed other models in terms of MAE and RMSE for each lookback, especially for larger lookback values (i.e., 12 and 16). Furthermore, the comparison with the MLP-Mixer method shows that the proposed method has a significant advantage in prediction accuracy, especially when preprocessing steps are taken into account. These findings suggest that the proposed LR-FFT-RP-MLP method has the potential to be a useful tool for wind energy forecasting applications.

\section{Conclusion}
\noindent In this paper, we propose a novel hybrid wind speed prediction model that utilizes a rank pooling and fast Fourier transform strategy to model the temporal evolution of wind speed. Our approach employs multilayer perceptron (MLP), long short-memory network (LSTM), and linear regression (LR) models as predictors. In the proposed hybrid FFT-RP-MLP model, we use RP and FFT to extract both local and global features from raw meteorological data, and then use MLP and LSTM to perform preliminary wind speed prediction.
To evaluate the performance of our proposed model, we conducted experiments on ten different settings, including the MLP model, LSTM model, FFT-MLP model, FFT-LSTM model, RP-MLP model, RP-LSTM model, FFT-RP-MLP model, FFT-RP-LSTM model, LR-FFT-RP-MLP model, and LR-FFT-RP-LSTM model. These models were evaluated based on real meteorological data, and our proposed hybrid model demonstrated strong capability in multistep wind speed prediction.
Our comparative analysis demonstrated that our proposed hybrid model outperformed several other similar models in terms of prediction accuracy and stability. Our results suggest that our proposed model represents a promising approach for wind speed prediction. This research offers a novel perspective for the improvement of advanced multistep wind speed prediction.
In conclusion, our findings indicate that the proposed hybrid model yields satisfactory performance in multistep wind speed forecasting, providing evidence of the effectiveness of our approach. This work has the potential to contribute to the field of renewable energy by improving the accuracy and reliability of wind speed predictions, which in turn can aid in the efficient planning and operation of wind energy systems.

While the proposed hybrid wind speed prediction model demonstrated promising results, it also has some limitations. Specifically, the feature extraction method and prediction model used are relatively basic and the parameter settings of the model are simplistic. Additionally, the prediction time step is not long enough to fully capture the complexity of wind speed temporal evolution.

Future work could explore the use of additional preprocessing techniques or alternative neural network architectures to further improve the accuracy of multistep wind speed prediction. Additionally, other meteorological parameters, such as temperature and humidity, could be integrated into the model to improve prediction accuracy. Finally, the proposed method could be tested on data from other regions to assess its generalizability and potential for wider application.
\\

Overall, this study provides a foundation for further development and improvement of the multistep advanced prediction of wind speed. The data that support the findings of this study are available from the corresponding author upon reasonable request, which will facilitate further research in this area.


%





\ifCLASSOPTIONcaptionsoff
  \newpage
\fi



%
%
%
\bibliographystyle{elsarticle-num}
\bibliography{ref}

%








\end{document}